\documentclass[letterpaper]{article} 
\usepackage{aaai23}  
\usepackage{times}  
\usepackage{helvet}  
\usepackage{courier}  
\usepackage[hyphens]{url}  
\usepackage{graphicx} 
\usepackage{natbib}  
\usepackage{caption} 
\usepackage{algorithm}
\usepackage{algorithmic}

\usepackage[utf8]{inputenc} 

\usepackage{appendix}
\usepackage{amsthm}
\usepackage{amsmath}

\usepackage{amsfonts}
\usepackage{multirow}
\usepackage{multicol}

\usepackage{color}

\usepackage{xcolor,colortbl}
\definecolor{LightCyan}{rgb}{0.88,1,1}

\usepackage{subfigure}
\usepackage{tcolorbox}

\newtheorem{theorem}{Theorem}
\newtheorem{lemma}{Lemma}

\newtheorem{assumption}{Assumption}
\newtheorem{remark}{Remark}

\usepackage{newfloat}
\usepackage{listings}
\DeclareCaptionStyle{ruled}{labelfont=normalfont,labelsep=colon,strut=off} 
\floatstyle{ruled}
\newfloat{listing}{tb}{lst}{}
\floatname{listing}{Listing}
\title{Decentralized Riemannian Algorithm for Nonconvex Minimax Problems}
\author {
    Xidong Wu,
    Zhengmian Hu,
    Heng Huang
}
\affiliations {
    Electrical and Computer Engineering, University of Pittsburgh, Pittsburgh, PA, United States \\
    xidong\_wu@outlook.com, 
    huzhengmian@gmail.com,
    henghuanghh@gmail.com
}

\usepackage{bibentry}

\begin{document}

\maketitle

\begin{abstract}
The minimax optimization over Riemannian manifolds (possibly nonconvex constraints) has been actively applied to solve many problems, such as robust dimensionality reduction and deep neural networks with orthogonal weights (Stiefel manifold). Although many optimization algorithms for minimax problems have been developed in the Euclidean setting, it is difficult to convert them into Riemannian cases, and algorithms for nonconvex minimax problems with nonconvex constraints are even rare. On the other hand, to address the big data challenges, decentralized (serverless) training techniques have recently been emerging since they can reduce communications overhead and avoid the bottleneck problem on the server node. Nonetheless, the algorithm for decentralized Riemannian minimax problems has not been studied. In this paper, we study the distributed nonconvex-strongly-concave minimax optimization problem over the Stiefel manifold and propose both deterministic and stochastic minimax methods. The Steifel manifold is a non-convex set. The global function is represented as the finite sum of local functions. For the deterministic setting, we propose DRGDA and prove that our deterministic method achieves a gradient complexity of $O( \epsilon^{-2})$ under mild conditions. For the stochastic setting, we propose DRSGDA and prove that our stochastic method achieves a gradient complexity of $O(\epsilon^{-4})$. The DRGDA and DRSGDA are the first algorithms for distributed minimax optimization with nonconvex constraints with exact convergence. Extensive experimental results on the Deep Neural Networks (DNNs) training over the Stiefel manifold demonstrate the efficiency of our algorithms.
\end{abstract}

\section{Introduction}
The minimax problems arise throughout machine learning, such as generative adversarial networks (GANs) \citep{goodfellow2014generative}, adversarial training \cite{wu2022retrievalguard}, and policy evaluation \citep{wai2019variance}. In recent years, minimax problems over Riemannian manifolds (possibly nonconvex constraints) have attracted intensive interest because
of their important applications in many scientific and engineering fields \citep{jordan2022first}, 
such as subspace robust Wasserstein distance \citep{paty2019subspace}, distributionally robust principal component analysis (PCA) \citep{zhang2022minimax}, adversarial training over Riemannian manifolds \citep{huang2020riemannian} and orthonormal generative adversarial networks (GANs) \citep{muller2019orthogonal}. Among them, optimization over the Stiefel manifold is popular because, from robust PCA to deep neural networks with orthogonal weights, a plethora of machine learning applications can be cast into it \citep{huang2018orthogonal}.

As sizes of model parameters and training datasets keep increasing, machine learning tasks heavily rely on distributed training \citep{bao2022doubly, lu2019gnsd, wu2022faster}. Given that computing capacity increases rapidly, communication overhead gradually becomes the bottleneck of training speed. As a result, research on communication efficient distributed optimization is very important. One important direction is 
decentralized distributed training, where each worker node updates the local model based on its local data and communicates exclusively with its neighbors, instead of a central server \citep{lian2017can}. 
In other words, when there are a lot of worker nodes, the decentralized distributed training strategy is more conducive to communication. 
In addition, due to data heterogeneity in the distributed setting, distributed robust training has wide applications. The distributed robustification of many ML tasks can also be formulated as decentralized minimax optimization problems \citep{jordan2022first}. 

Although decentralized distributed training has gained popularity, algorithms for decentralized Riemannian minimax problems have not been studied. The existing minimax optimization methods over Riemannian manifolds cannot be applied to these problems \citep{jordan2022first, zhang2022minimax,han2022riemannian,huang2020riemannian}, since (1) they mainly focus on the geodesically strongly-convex-concave case or require PL condition, which might not be satisfied in the deep learning and nonconvex optimization \citep{han2022riemannian}; (2) algorithms cannot be used in the distributed setting. 

In this paper, we study the following distributed nonconvex-strongly-concave minimax problem on Stifel manifolds:
\begin{align} \label{eq:1}
& \min_{x \in \mathrm{St}(d,r)} \max_{y \in \mathcal{Y}} F(x,y) = \frac{1}{n} \sum_{i=1}^{n} f_{i} \left(x_i, y_i\right) \\
 & \text { s.t. } \quad x_{i}=x_{j}, y_{i}=y_{j}, (i,j) \in \mathcal{E} \nonumber
\end{align}
where the function $f_{i}(x,y): \mathcal{M} \times \mathcal{Y} \rightarrow \mathbb{R}$ is nonconvex over $x \in \mathcal{M}$ and is $\mu$-strongly concave over $y \in \mathcal{Y} \subseteq \mathbb{R}^{d}$ 
Here $ \mathcal{Y}\subseteq \mathbb{R}^{d}$ is nonempty compact convex sets but $\mathcal{M}:= \text{St}(d,r) = \{ x \in \mathbb{R}^{d \times r} :x^{\top}x = I_r \}$ is the Stiefel manifold. Unlike the Euclidean distributed setting, the Stiefel manifold is a non-convex set. $\mathcal{E}$ is the communication link set in the decentralized communication network. We consider a multi-client network system $\mathcal{G} = (\mathcal{V},  \mathcal{E}) $, where $\mathcal{V}$ = $\{1, \dots, n\}$ is the collection of work nodes indices and $\mathcal{E} \subseteq \mathcal{V} \times \mathcal{V}$ is the collection of communication links (i, j), i, j $\in \mathcal{V}$, that connect worker nodes. Each worker node i possesses local data and updates the local model.

In this paper, we also focus on the stochastic form of decentralized minimax problem \eqref{alg:1}, defined as
\begin{align} \label{eq:2}
& \min_{x \in \mathrm{St}(d,r) } \max_{y \in \mathcal{Y}} F(x,y) =  \frac{1}{n}  \sum_{i=1}^{n} \mathbb{E}[f_{i}(x,y;\xi)] \\
 & \text { s.t. } \quad x_{i}=x_{j}, y_{i}=y_{j}, (i,j) \in \mathcal{E} \nonumber
\end{align}
where $\xi$ is a random variable following an unknown distribution $\mathcal{D}_i$. Many important applications can be written in the form of \eqref{eq:1} and \eqref{eq:2}, such as robust geometry-aware PCA \citep{zhang2022minimax}, distributionally robust DNN training with orthogonal weights \citep{huang2020riemannian}. PCA has a wide range of applications, and it can be used 
K-means clustering \citep{zhang2022minimax} and study the functional diversity of cells \citep{baden2016functional}. For deep learning, orthonormality on parameters has been proven to increase generalization precision and speed up and stabilize convergence of neural network models \citep{wang2020orthogonal,huang2018orthogonal},
which corresponds to optimization over the Stiefel manifold. 
Although 
above minimax problems \eqref{eq:1} and \eqref{eq:2} have been frequently appeared in many machine learning applications, \eqref{eq:1} and \eqref{eq:2} in distributed settings have not been theoretically studied. The existing explicitly decentralized minimax optimization methods such as the decentralized gradient descent ascent method only focus on the minimax problems in Euclidean space. 

The change from Euclidean to Riemannian nonconvex optimization and from centralized to distributed settings pose several difficulties. For example, we cannot use the arithmetic mean as a consensus step on the Stifel manifold because the average result might not be on the Stifel manifold. Under the Riemannian PL condition or g-convex-concave \citep{jordan2022first, han2022riemannian}, variables x and y can be embedded as $z = (x^{\top}, y^{\top})$ and solve two variables together, which cannot be applied in our nonconvex case.  Then there exists a natural question:
\begin{center}
\begin{tcolorbox}
\vspace{-0.05in}
\textbf{ Can we design new decentralized algorithms with exact convergence for distributed minimax optimization on the Stiefel manifold for solving the Problems \eqref{eq:1} and  \eqref{eq:2}? }
\vspace{-0.05in}
\end{tcolorbox}
\end{center}
In the paper, we give an affirmative answer to the above question and propose a class of gradient descent ascent methods for solving the Problems \eqref{eq:1} and \eqref{eq:2}.
Our main contributions are three-fold:
\begin{itemize}
\item[1)] We propose the first decentralized gradient descent ascent algorithm (DRGDA) over the Stiefel manifold for the deterministic minimax problem \eqref{eq:1}. Our method is retraction-based and vector transport-free. We prove that the DRGDA has a sample complexity of $O( \epsilon^{-2})$ for finding an $\epsilon$-stationary point, which matches the best complexity result achieved by the centralized minimax method on Euclidean space.
\item[2)]  We also propose a decentralized stochastic first-order algorithm (DRSGDA) over the Stiefel manifold for the stochastic problem \eqref{eq:2}, which has a sample complexity of $O(\epsilon^{-4})$, which is the first theoretical result of the stochastic method in the decentralized Riemannian minimax problems.
\item[3)] Extensive experimental results on the robust DNN training over the Stiefel manifold confirm the effectiveness of our proposed algorithm.
\end{itemize}

\section{Related Works}

\subsection{Minimax Optimization and Decentralized Distributed Learning}
The minimax optimization recently has been successfully applied in many machine learning problems such as robust federated learning, reinforcement learning \cite{wai2019variance}, and adversarial training \cite{wu2022adversarial}. 
Many gradient-based minimax methods were proposed for solving these minimax optimization problems  \citep{lin2020gradient,lin2020near}. In addition, some works developed accelerated gradient descent ascent algorithms with the variance-reduced techniques \citep{luo2020stochastic}. 
More recently, \citet{huang2021efficient} proposed mirror descent ascent methods to solve the nonsmooth nonconvex-strongly-concave minimax problems based on dynamic mirror functions. \citet{huang2023adagda} used momentum-based variance-reduced technique to design the single-loop algorithm (Acc-MDA). 

At the same time, many decentralized distributed methods are proposed because they are useful when the network has communication bottlenecks or limited bandwidth between certain nodes and the central node. \citet{lian2017can} firstly provides theoretical support for the potential advantage of a decentralized distributed algorithm. Subsequently, D-SPIDER-SFO \cite{pan2020d}, DEEPSTORM \cite{mancino2022proximal} and many other methods are proposed for the decentralized training.

Recently, \citet{xian2021faster} and \citet{zhang2021taming} extended the minimax problem to a decentralized distributed setting and solve the nonconvex-strongly-concave minimax problems in the Euclidean spaces. \citet{zhang2021taming} proposed both deterministic gradient-based decentralized minimax methods and stochastic gradient-based decentralized minimax methods. 

\subsection{Riemannian Manifold Optimization }
Riemannian manifold optimization is an important problem given that they are related to many machine learning problems, such as low-rank matrix completion 
, dictionary learning
and deep learning  \citep{huang2018orthogonal}. \citet{liu2017accelerated} proposed first-order gradient methods for geodesically convex functions. Furthermore, \citet{zhang2016riemannian} applied variance-reduced technology to Riemannian manifold optimization and proposed fast stochastic methods. More recently, various fast Riemannian gradient-based methods have been proposed for non-convex optimization based on the retraction and vector transport \citep{kasai2018riemannian}. Then some efficient Riemannian adaptive optimization algorithms are introduced \citep{kasai2019riemannian}. 

Recently, \citet{huang2020riemannian} proposed a class of Riemannian gradient descent ascent algorithms to solve the nonconvex-strongly-concave minimax problems on Riemannian manifolds. Then \citet{han2022riemannian} introduced Riemannian Hamiltonian function and proposed Riemannian Hamiltonian gradient methods under the Riemannian PL condition. \citet{zhang2022minimax} and \citet{jordan2022first} proved that the Riemannian corrected extragradient (RCEG) method achieves convergence at a linear rate in the geodesically strongly-convex-concave problems. In addition, some methods are proposed for variational inequalities, which are the implicit minimax problems, on Riemannian manifolds \citep{huang2020riemannian}.

In addition, several algorithms were proposed for distributed optimization on Riemannian manifolds \citep{chen2021decentralized}. In the decentralized distributed setting, the consensus step is usually done by calculating the Karcher mean on the manifold \citep{shah2017distributed}, or calculating the minimizer of the sum of the square of the Euclidean distances in the corresponding sub-manifold case \citep{chen2021decentralized}. \citet{grammenos2020federated} firstly present the algorithm for federated learning on Riemannian manifolds and focus on the PCA problems (Stiefel manifold). 

\section*{Notations}
We use $\mathbf{x}$ and $\mathbf{y}$ to denote a collection of all local parameters $x_i, y_i$, i.e., $\mathbf{x}^{\top} = (x_{1}^{\top}, x_{2}^{\top}, \dots, x_{N}^{\top})$ and $\mathbf{y}^{\top} = (y_{1}^{\top}, y_{2}^{\top}, \dots, y_{N}^{\top})$. $\otimes$ denotes Kronecker product. Define $\bar{\mathbf{x}} = \mathbf{1} \otimes \bar{x}$.  $\|\cdot\|$ denotes the $\ell_2$ norm for vectors and Frobenius norm for matrices, respectively.
$\nabla_x f(x,y)$ and $\nabla_y f(x,y)$
denote the Euclidean partial derivatives w.r.t. variables $x$ and $y$ respectively, and the Riemannian gradient is $\mathrm{grad}_x g(x, y)$. For $\mathcal{M}$, the $n-$fold Cartesian product of $\mathcal{M}$ is denoted as $\mathcal{M}_n = \mathcal{M} \times \dots \times \mathcal{M}$. $I_r$ denotes $r$-dimension identity matrix. 
Given the mini-batch samples $\mathcal{B}=\{\xi_i\}_{i=1}^q$, we let $\nabla f(x,y;\mathcal{B})=\frac{1}{q}\sum_{i=1}^q \nabla f(x,y;\xi_i)$. 
In general, the manifold $\mathcal{M}$ is endowed with a smooth inner product $\langle x,y\rangle_x : \mathrm{T}_x \mathcal{M} \times \mathrm{T}_x \mathcal{M} \rightarrow R$ on tangent space $\mathrm{T}_x \mathcal{M}$ for every $x \in \mathcal{M}$. 

The network system $\mathcal{G} = (\mathcal{V},  \mathcal{E}) $ is represented by double stochastic matrix $W$, which is defined as follows: (1) if there exists a link between node i and node j, then $\mathbf{W}_{ij} > 0$, otherwise $\mathbf{W}_{ij}$ = 0, (2) $\mathbf{W} = \mathbf{W}^{\top}$ and (3) $\mathbf{W1} = \mathbf{1}$ and $\mathbf{1}^{\top} \mathbf{W} = \mathbf{1}^{\top}$. The second-largest eigenvalue of $\mathbf{W}$ as $\lambda$.

\section{Preliminaries}

\subsection{Riemannian Gradients} Given that the metric on tangent space $\mathrm{T}_x \mathcal{M}$ is induced from the Euclidean inner product $ \langle \cdot, \cdot \rangle$, the Riemannian gradient grad$h(x)$ on $\mathrm{St}(d,r)$ is defined as  $\mathrm{grad}h(x) = \mathcal{P}_{\mathrm{T}_{x} \mathcal{M}} \nabla h(x)$, where $\mathcal{P}_{\mathrm{T}_{x} \mathcal{M}}$ is the orthogonal projection onto $\mathrm{T}_{x} \mathcal{M}$. On the $\mathrm{St}(d,r)$, we have
\begin{align}
\mathcal{P}_{\mathrm{T}_{x} \mathcal{M}} y = y - \frac{1}{2}x(x^{\top}y + y^{\top}x)
\end{align}


\subsection{Retraction}
Given $x \in \mathcal{M}$ and $u \in \mathrm{T}_x \mathcal{M}$, retraction $R_x: \mathrm{T}_x \mathcal{M} \rightarrow \mathcal{M}$ is defined as mapping from tangent space $\mathrm{T}_x \mathcal{M}$ onto $\mathcal{M}$ with a local rigidity condition that preserves the gradients at $x \in \mathcal{M}$. The retraction $R_x$ satisfies: 
1) $R_x(0)=x$. 2) $D R_x(0)=id_{T_x \mathcal{M}}$.

In the centralized Riemannian gradient descent iteration, the update step is defined as \begin{align}
x_{t+1} = R_{x_t}( - \alpha  \mathrm{grad} h(x_t)),
 \end{align}
namely, it updates parameters along a negative Riemannian gradient direction on the tangent space, then it uses an operation called retraction $R_{x_t}$ to map onto $\mathcal{M}$. Following  \citep{chen2021decentralized}, we also assume the second-order boundedness of retraction $R_{x}$, namely,  
\begin{align}
R_x(u) = x + u + O(\|u \|^2).
\end{align}

Furthermore, following \citep{boumal2019global,chen2021decentralized}, we have the following lemma:
\begin{lemma} \citep{boumal2019global} \label{lem:1}
Let $R$ be a second-order retraction over $\mathrm{St}(d,r)$, we have
\begin{equation}
\begin{aligned}
\|R_x(u) - (x + u)\| &\leq M \|u \|^2 \\
\forall x \in \mathrm{St}(d,r),& \forall u \in \mathrm{T}_x \mathcal{M}
\end{aligned}
\end{equation}
Moreover, if the retraction is the polar decomposition. For all $x \in \mathrm{St}(d,r)$ and $u \in \mathrm{T}_x \mathcal{M}$, the following inequality holds :
\begin{equation}
\begin{aligned}
\|R_x(u) - z\| \leq \|x + u - z\|, \forall z \in \mathrm{St}(d,r)
\end{aligned}
\end{equation}
\end{lemma}
Unless otherwise stated, the sequel uses the polar retraction to offer a concise analysis. More details on the polar retraction are provided in  \citep{liu2019quadratic}. 

\subsection{Mean and Consensus on the Stiefel Manifold}
In the Euclidean decentralized distributed algorithms, the $\|x_i - \bar{x}\|^2$ is usually used as the consensus error, where $\bar{x}$ is defined as the Euclidean average point of $x_1, \cdots,  x_n$ by
\begin{align} \label{eq:8}
\bar{x} = \frac{1}{n} \sum_{i=1}^{n} x_i
\end{align}
However, in the decentralized distributed Riemannian setting, such as $\mathrm{St}(d,r)$, we cannot directly average local parameters as in \eqref{eq:8} because the arithmetic mean result might not be on the manifold. Therefore, we follow  \citep{sarlette2009consensus} and use induced arithmetic mean (IAM) on $\mathrm{St}(d,r)$, defined as 
\begin{align}
\hat{x}:=\underset{z \in \operatorname{St}(d, r)}{\operatorname{argmin}} \sum_{i=1}^{n}\left\|z-x_{i}\right\|
^{2}=\mathcal{P}_{\mathrm{St}}(\bar{x})  
\end{align}
where $\mathcal{P}_{\mathrm{St}}(\cdot)$ is the orthogonal projection onto $\mathrm{St}(d,r)$. And similarly we have 
\begin{align}
\underset{z \in \operatorname{St}(d, r)}{\operatorname{min}} \sum_{i=1}^{n}\frac{1}{n} \left\|z-x_{i}\right\|^{2} = \frac{1}{n} \|\mathbf{x} - \hat{\mathbf{x}} \|^2     
\end{align}
Furthermore, we also define 
\begin{align}
\|\mathbf{x} - \hat{\mathbf{x}}\|_{\infty} = 
\underset{i \in [n]}{\mathrm{max}} \left\|x_{i} - \hat{x} \right\|    
\end{align}

As for the consensus problem over the Stiefel manifold, following the \citep{chen2021decentralized}, we use the minimizer of the sum of the square of the $l_2$ norm over $\mathrm{St}(d,r)$. It means:
\begin{align} \label{eq:12}
\min J^{k}(\mathbf{x}) &:=\frac{1}{2} \sum_{i=1}^{n} \sum_{j=1}^{n} W_{i j}^{k}\left\|x_{i} - x_{j}\right\|^{2} \\
\text { s.t. } \quad x_{i} & \in \mathcal{M}, \forall i \in[n] \nonumber
\end{align}
where the superscript $k \geq 1 $ is an integer. It should be mentioned we use $k$ to offer flexibility in algorithm design and analysis. And $\mathbf{W}_{ij}^{k}$ corresponds to performing $k$ steps of communication on the tangent space. 


\section{ Decentralized Riemannian Gradient Descent Ascent Methods }

In the section, we propose deterministic and stochastic methods for solving decentralized Riemannian minimax Problems in \eqref{eq:1} and \eqref{eq:2}, respectively. 

\subsection{DRGDA Algorithms}
In this subsection, we first study the decentralized Riemannian gradient descent ascent (DRGDA) method with gradient tracking technologies for solving the problem \eqref{eq:1}. We describe our algorithm in Algorithm \ref{alg:1}. 

\begin{algorithm}[tb]
\caption{ DRGDA Algorithm }
\label{alg:1}
\begin{algorithmic}[1] 
\STATE {\bfseries Input:} $T$, tuning parameters $\{\alpha, \beta, \eta \}$; \\
\STATE {\bfseries initialize:} Initial input $x_0^i \in \mathrm{St}(d,r)$, $y_0^i \in \mathcal{Y}$,
and then compute $u_0^i=\mathrm{grad}_x f_i(x_0^i,y_0^i)$ and $v_0^i = \nabla_y f(x_0^i,y_0^i)$, where $i \in [n]$; \\
\FOR{$t = 0, 1, \ldots, T$}
\STATE $x_{t+1}^i = R_{x_t^i}( \mathcal{P}_{\mathrm{T}_{x_t^i}\mathcal{M}}(\alpha\sum_{j=1}^{n}\mathbf{W}_{ij}^k x_t^j) - \beta w_t^i)$  where $w_t^i = \mathcal{P}_{\mathrm{T}_{x_t^i}\mathcal{M}}(u_t^i)$ \\
\STATE $y_{t+1}^i = \sum_{j=1}^{n}\mathbf{W}_{ij}^k y_t^{j} + \eta v_t^{j}$; \\
\STATE $u_{t+1}^i =\sum_{j=1}^{n} \mathbf{W}_{ij}^k  u_{t}^j + \mathrm{grad}_x f_{i}\left(x_{ t+1}^i, y_{t+1}^i\right)-\mathrm{grad}_x f_{i}\left(x_t^i,y_t^i\right) $;
\STATE $v_{t+1}^i =\sum_{j=1}^{n} \mathbf{W}_{ij} v_{t}^j + \nabla_y f_{i}\left(x_{ t+1}^i, y_{t+1}^i\right)- \nabla_y f_{i}\left(x_t^i, y_t^i\right) $;
\ENDFOR
\end{algorithmic}
\end{algorithm}

At the beginning of Algorithm \ref{alg:1}, one can simply initialize local model parameters $x \in \mathcal{M}$ and $y \in \mathcal{Y}$ for all worker nodes with the same points. 

At the step 4 of Algorithm \ref{alg:1}, we use the decentralized Riemannian gradient method to update variable x based on the gradient tracker $u_t^i$. We firstly need to project the gradient tracker $u_t^i$ onto the tangent space $\mathrm{T}_{x_t^i} \mathcal{M}$, which follows a retraction consensus update (Seen in the supplementary materials).
We can regard the step 4 as applying Riemannian gradient method to solve the following problem:
\begin{align}
\min _{\mathbf{x} \in \mathrm{St}(d,r)} \beta F(\mathbf{x})+\alpha J^{k}(\mathbf{x})
\end{align}
where $J^{k}(\mathbf{x})$ is the consensus problem defined in \eqref{eq:12} and more details seen in the supplementary materials.
The consensus step and computation of Riemannian gradient can be done in parallel \citep{chen2021decentralized}.

At the step 5 of Algorithm \ref{alg:1}, we follow conventional decentralized gradient ascent steps, where we update local parameters with gradient tracker and then do the consensus step with double
stochastic matrix $\mathbf{W}$.

At the step 6 of Algorithm \ref{alg:1}, we use $u_t^i$ to tracks the average Riemannian gradient $\mathrm{grad}_x f_i(x_t^i,y_t^i)$. 
Gradient tracking is a popular technology in the decentralized distributed learning to
guarantee the convergence of the global objective function. Note, we can regard $\mathrm{grad}_x f_i(x_t^i,y_t^i)$ at the step 6 as the projected Euclidean gradient, and we do not need to project it on the tangent space $\mathrm{T}_{x_t^i} \mathcal{M}$ to save the computation cost, and we only do the projection at the step 4. 

At the step 7 of Algorithm \ref{alg:1}, we use the similar operation.
Each worker node calculates the local full gradients, and then we update the gradient tracker $v_t^i$ for y.

\subsection{DRSGDA Algorithms}
We also consider stochastic method because Stochastic gradient descent are widely used in learning tasks \cite{sun2022demystify}. In the DRGDA algorithm, worker nodes need to compute local full gradients in each iteration, which result in a high sample complexity when the local datasets are large. It also cannot be applied in the online learning. This limitation motivates us to leverage the stochastic method and propose Riemannian decentralized stochastic gradient descent ascent (DRSGDA) method. Algorithm \ref{alg:2} show the algorithmic framework of DRSGDA method. 
 
 At the step 4 and step 5 of Algorithm \ref{alg:2}, we use the similar step to update variable $ \mathrm{St}(d,r)$ and $y \in \mathcal{Y}$. The different is that we use the $\underline{u}_t$ and $\underline{v}_t$, which is the gradient tracker for the unbiased gradient estimators.
At the steps 7 and 8 of Algorithm \ref{alg:2}, we calculate the stochastic Riemannian gradient for variable x, and calculate the stochastic gradient for variable y based on sampled data to update the $\underline{u}_t$ and $\underline{v}_t$.

\begin{algorithm}[tb]
\caption{ DRSGDA Algorithm }
\label{alg:2}
\begin{algorithmic}[1] 
\STATE {\bfseries Input:} $T$, tuning parameters $\{\alpha, \beta, \eta \}$; \\
\STATE {\bfseries initialize:} Initial input $x_0^i \in \mathrm{St}(d,r)$, $y_0^i \in \mathcal{Y}$,
and then compute $u_0^i=\mathrm{grad}_x f_i(x_0^i,y_0^i; \mathcal{B}_{0}^i)$ and $v_0^i = \nabla_y f(x_0^i,y_0^i; \mathcal{B}_{0}^i)$, where $i \in [n]$; \\
\FOR{$t = 0, 1, \ldots, T$}
\STATE $x_{t+1}^i = R_{x_t^i}( \mathcal{P}_{\mathrm{T}_{x_t^i}\mathcal{M}}(\alpha\sum_{j=1}^{n}\mathbf{W}_{ij}^k x_t^j) - \beta w_t^i)$  where $w_t^i = \mathcal{P}_{\mathrm{T}_{x_t^i}\mathcal{M}}(\underline{u}_t^i)$ \\
\STATE $y_{t+1}^i = \sum_{j=1}^{n}\mathbf{W}_{ij}^k y_t^{j} + \eta v_t^{j}$
\STATE Draw a mini-batch i.i.d. samples $\mathcal{B}_{t+1}^i=\{\xi_{j}^i\}_{j=1}^q$, and then compute \\ 
\STATE $\underline{u}_{t+1}^i =\sum_{j=1}^{n} \mathbf{W}_{ij}^k \underline{u}_{t}^j + \mathrm{grad}_x f_{i}\left(x_{ t+1}^i, y_{t+1}^i; \mathcal{B}_{t+1}^i\right)$ \\ 
$- \mathrm{grad}_x f_{i}\left(x_t^i,y_t^i; \mathcal{B}_{t}^i\right)$;\\
\STATE $\underline{v}_{t+1}^i =\sum_{j=1}^{n} \mathbf{W}_{ij} \underline{v}_{t}^j + \nabla_y f_{i}\left(x_{ t+1}^i, y_{t+1}^i; \mathcal{B}_{t+1}^i\right) - \nabla_y f_{i}\left(x_t^i, y_t^i; \mathcal{B}_{t}^i\right) $;
\ENDFOR
\end{algorithmic}
\end{algorithm}

\section{Convergence Analysis}
\subsection{Assumptions }
In the subsection, we state assumptions for  the problems.

\begin{assumption} \label{ass:1}
Each component function $f_i(x,y)$ is twice continuously differentiable in both $x \in \mathrm{St}(d,r)$ and $y \in \mathcal{Y}$, 
For $i \in [n]$, we have
\begin{align}
& \|\nabla_x f_i(x_1,y)-\nabla_x f_i(x_2,y)\| \leq L_{11} \|x_1 - x_2\|, \nonumber \\
& \|\nabla_x f_i(x,y_1)-\nabla_x f_i(x,y_2)\| \leq L_{12} \|y_1 - y_2\|, \nonumber \\
& \|\nabla_y f_i(x_1,y)-\nabla_y f_i(x_2,y)\| \leq L_{21} \|x_1 - x_2\|, \nonumber \\
& \|\nabla_y f_i(x,y_1)-\nabla_y f_i(x,y_2)\| \leq L_{22} \|y_1 - y_2\|.  \nonumber
\end{align}
and we also assume the uniform upper bound of $ \|\mathrm{grad}_x f_i (x_t^i, y_t^i) \|$ is D.
\end{assumption}
 
The $L$-smooth assumption is widely used in the minimax problems in the Euclidean space. Since the Stiefel manifold is embedded in Euclidean space, we are free to do so, and detailed comparison is provided by \citep{chen2021decentralized}. 

One could also consider the following Lipschitz inequality \citep{huang2020riemannian},
\begin{align}
\left\|\mathrm{P}_{x \rightarrow y} \operatorname{grad} f(x)-\operatorname{grad} f(y)\right\|_{\mathrm{F}} \leq L_{g}^{\prime} d_{g}(x, y)
\end{align}
Because it involves vector transport and geodesic distance, which add computational burden, we choose to use  assumption \ref{ass:1} for simplicity. And $L_{11}, L_{12}, L_{21}, L_{22}$ are all constants and we set $L = \max\{L_{11}, L_{12}, L_{21}, L_{22}\}$. Since global function $F(x, y)$ is a finite sum of local functions, it is also the L-smooth. 

We also bound the partial derivative with respect to variable $x$ as \citep{chen2021decentralized}. It is reasonable since the Stiefel manifold is compact and often satisfied in practice. One common example is the finite-sum form of function.

\begin{assumption} \label{ass:2}
The function $f_i(x,y)$ is $\mu$-strongly concave in $y\in \mathcal{Y}$, i.e., for all $x\in \mathrm{St}(d,r)$ and $y_1, y_2\in \mathcal{Y}$,
then the following inequality holds
\begin{align}
 f_i(x,y_1) \leq& f_i(x,y_2) + \langle\nabla_y f_i(x,y_2), y_1-y_2\rangle \nonumber\\
 &- \frac{\mu}{2}\|y_1-y_2\|^2. 
\end{align}
\end{assumption}

Then we could easily get $F(x,y)$ is also $\mu$-strongly in $y\in \mathcal{Y}$.
Then, there exists a unique solution to
the problem $\max_{y\in \mathcal{Y}} F(x,y)$ for any $x$. Here we let $y^*(x) = \arg\max_{y \in \mathcal{Y}} F(x,y)$
and $\Phi(x) = F(x,y^*(x)) = \max_{y\in \mathcal{Y}} F(x,y)$.

\begin{assumption} \label{ass:3}
The function $\Phi(x)$ is bounded below in $\mathrm{St}(d,r)$, \emph{i.e.,} $\Phi^* = \inf_{x\in \mathrm{St}(d,r)}\Phi(x) > -\infty$.
\end{assumption}

\subsection{A Useful Convergence Metric }
Since finding the global optimal solution for the non-convex min-max problem is NP-hard in general \citep{lin2020gradient},  we firstly introduce a useful convergence metric to measure convergence of
our algorithms for both deterministic and stochastic settings. Given the variable sequence  generated from our algorithms, we
define a convergence metric as follows:
\begin{align}
\mathfrak{M}_t & =\| \mathrm{grad}_x F(x_t,y_t)\| +  \frac{1}{n}\|x_t - \hat{x}_{t}\|  \nonumber \\
& \quad + \frac{L}{n} \|\bar{y}_t - y^*(\hat{x}_t)\|,
\end{align}
where the first and second terms of $\mathfrak{M}_t$ measure the optimality gap of non-convex minimax problems, while the last term measures the convergence of $\bar{y}_t$ to the unique maximizer  of $F(\hat{x}_t, \cdot)$

\subsection{Convergence Analysis of DRGDA Algorithms}
In the subsection, we study the convergence properties of our DRGDA algorithm. 

\begin{lemma} \label{lem:2}
Under the above assumptions, the gradient of function $\Phi(x) = \max_{y\in \mathcal{Y}} F(x, y)$ is $L_{\Phi}$-Lipschitz with respect to retraction, and the mapping or function $y^{*}(x) = \arg \max_{y \in \mathcal{Y}} f(x, y) $ is $\kappa$-Lipschitz with respect to retraction. $\forall x_1, x_2 \in \mathcal{M} $ and $u \in \mathrm{T}_{x_1} \mathcal{M}$, we have
\begin{align}
\|y^{*}(x_1) - y^{*}(x_2) \| &\leq \kappa \|x_1 - x_2\| \nonumber\\
\| \mathrm{grad}\Phi(x_1) - \mathrm{grad}\Phi(x_2) \| &\leq L_{\Phi} \|x_1 - x_2\|  \nonumber
\end{align}
where $ L_{\Phi} = \kappa L_{12} + L_{11}$,and $\kappa = L_{21} / \mu$ denotes the number condition.
\end{lemma}

The lemma \ref{lem:2} guarantees that $\Phi(x)$ is $L_{\Phi}$-Lipschitz and $y^*(\cdot)$ is $\kappa$-Lipschitz, and it provides the theoretical support to use $\Phi(x)$ in the proof.

\begin{theorem} \label{thm:1}
Under Assumptions (\ref{ass:1}, \ref{ass:2} and \ref{ass:3}) Let $k \geq \lceil log_{\lambda_2} ( \frac{1}{2\sqrt{n}} ) \rceil$, $\alpha \leq \frac{1}{M}$, 
$0 < \beta \leq \min \left\{\beta_0, \frac{1}{L + 9 C_2 }, \frac{\alpha \delta_{1}}{10 D}\right\}$ and $\eta \leq \frac{1}{L}$. And suppose the sequence  generated from Algorithm \ref{alg:1} and the initial parameters start from the same points, we have 
\begin{align}
 \frac{1}{T} \sum_{t=0}^{T-1}  \mathfrak{M}_t 
\leq&\frac{10( H_0 - H_{T} + C_{4})}{T \beta}  \nonumber\\
&+ \frac{10 \beta ( H_0 - H_{T} + C_{4}) C_3}{T} 
\end{align}
where $\beta_0$ and $Q$ is introduced in the proof of Theorem 1, and we let
\begin{align}
H_t = &\Phi(\hat{x}_t) + \frac{8L}{\mu \eta} \left\|\bar{y}_{t+1}-y^{*}\left(\hat{x}_{t+1}\right)\right\|^{2} + \frac{ Q}{n} \|\mathbf{y}_t - \bar{\mathbf{y}}_t\|^2 \nonumber\\
&+ \frac{\eta Q}{n} \|\mathbf{v}_t - \bar{\mathbf{v}}_t \|^2 \nonumber
\end{align}
and $C_{1} = \frac{1}{\left(1-\rho_{k}\right)^{2}}$; $C_2 = [\frac{1}{2 \eta} + \frac{  L_{\Phi}}{2} + \frac{40 \kappa^{2}L}{ \mu^2 \eta^2 } + (1 + \frac{1}{c_0}) 6L^2 \eta Q]$; $C_3 = \frac{2}{\left(1-\rho_{k}\right)^{2}}$ $C_4=\frac{C_3 C_7 \delta_{1}^{2} \alpha^{2}}{10} $; $C_5 = \left(2 M D+L + C_2( 5 \delta_{1}^2 + 5 )\right); C_6 = \frac{100 r D^{2} C_{1}}{L} + 2 \eta L^2 + \frac{67 L^2}{\mu^2} + 6L^2 \eta Q + C_2 101 \alpha^{4} C_1 \delta_1^2 + 8 M D \alpha^{2}+2 D \alpha + L; C_7 = 2L^2 \eta Q + \frac{100 r D^{2} C_{1}}{L}]$; $C_8 = \frac{2}{\left(1-\sigma_{2}^{t}\right)^{2}}$
\end{theorem}

\begin{remark}
To achieve an $\epsilon^2$-stationary solution, i.e., $ \frac{1}{T} \sum_{t=0}^{T-1}  \mathfrak{M}_t 
\leq \epsilon^2$, the total samples evaluated across the network system is on the order of $O(m \epsilon^{-2})$, where $m$ is the total number of data points.
\end{remark}

\subsection{Convergence Analysis of DRSGDA Algorithms}

\begin{assumption} \label{ass:4}
Each component function $f_i(x,y;\xi)$ has an unbiased stochastic gradient with
bounded variance $\sigma^2$, i.e.,  $\forall i \in [n], x \in \mathrm{St}(d,r), y \in \mathcal{Y}$
\begin{align}
& \mathbb{E}[\nabla f_i(x,y;\xi)] = \nabla f_i(x,y), \nonumber \\
& \mathbb{E}\|\mathrm{grad}_x f_i(x,y;\xi) - \mathrm{grad}_x f_i(x,y)\|^2 \leq \sigma_1^2 \nonumber\\
& \mathbb{E}\|\nabla_y f_i(x,y;\xi) - \nabla_y f_i(x,y)\|^2 \leq \sigma_2^2. \nonumber
\end{align}
and we define $\sigma = \max \{\sigma_{1}, \sigma_{2}\}$. In addition, we also assume a uniform upper bound of $ \|\mathrm{grad}_x f_i (x_t^i, y_t^i; \mathcal{B})\|$ is D.
\end{assumption}

Assumption \ref{ass:4} imposes the bounded variance of stochastic (Riemannian) gradients, which is commonly used in the stochastic optimization \citep{lin2020gradient,huang2020riemannian}.

\begin{theorem} \label{thm:2}
Under Assumptions (\ref{ass:1}, \ref{ass:2}, \ref{ass:3} and \ref{ass:4}) Let $k \geq \lceil log_{\lambda_2} ( \frac{1}{2\sqrt{n}} ) \rceil$, $\alpha \leq \frac{1}{M}$, 
$0 < \beta \leq \min \{\beta_0, \frac{1}{L + 9 C_2 }, \frac{\alpha \delta_{1}}{2(\tilde{\gamma}\sigma + 5D)}\}$ and $\eta \leq \frac{1}{L}$. And suppose the sequence  generated from Algorithm \ref{alg:2} 
\begin{align}
& \frac{1}{T} \sum_{t=0}^{T-1} \mathbb{E} \mathfrak{M}_t 
\leq 10\frac{\mathbb{E}H_{0} - \mathbb{E}H_T + C_{4}}{T \beta} \nonumber\\
&+ \frac{20 \beta( \mathbb{E}H_0 - \mathbb{E}H_{T} + C_{4}) C_3 }{T} + \frac{C_9 \sigma^2}{B}
\end{align}
where $\beta_0$ and $Q$ is introduced in the proof of Theorem 2 and we let 
\begin{align}
H_t = &\Phi(\hat{x}_t) + \frac{8L}{\mu \eta} \left\|\bar{y}_{t+1}-y^{*}\left(\hat{x}_{t+1}\right)\right\|^{2} + \frac{ Q}{n} \|\mathbf{y}_t - \bar{\mathbf{y}}_t\|^2 \nonumber\\
&+ \frac{\eta Q}{n} \|\mathbf{\underline{v}}_t - \bar{\mathbf{\underline{v}}}_t \|^2 \nonumber
\end{align}
and $C_{1} = \frac{1}{\left(1-\rho_{k}\right)^{2}}$; $C_2 = [\frac{1}{2 \eta} + \frac{  L_{\Phi}}{2} + \frac{40 \kappa^{2}L}{ \mu^2 \eta^2 } + (1 + \frac{1}{c_0}) 18 L^2 \eta Q]$; $C_3 = \frac{2}{\left(1-\rho_{k}\right)^{2}}$ $C_4=\frac{C_3 C_7 \delta_{1}^{2} \alpha^{2}}{4} $; $C_5 = \left(2 M D+L + C_2( 5 \delta_{1}^2 + 7)\right); C_6 = \frac{100 r D^{2} C_{1}}{L} + 2 \eta L^2 + \frac{67 L^2}{\mu^2} + 6L^2 \eta Q + C_2 101 \alpha^{4} C_1 \delta_1^2 + 8 M D \alpha^{2}+2 D \alpha + L; C_7 = 2L^2 \eta Q + \frac{100 r D^{2} C_{1}}{L}]$; $C_8 = \frac{2}{\left(1-\sigma_{2}^{t}\right)^{2}}, C_9 = [2 C_3 + 4(C_5 + C_6 C_3 + C_7 C_3) + (48 \alpha^{2} C_3 C_8 + 12 C_8)L^2] \beta \tilde{\gamma} +  180 \eta Q(1 + \frac{1}{c_0}) \frac{\sigma^2}{B \beta} + \frac{195 C_2 \beta \tilde{\gamma}^2\sigma^2}{B} $
\end{theorem}
\begin{remark}
To achieve an $\epsilon^2$-stationary solution, i.e., $ \frac{1}{T} \sum_{t=0}^{T-1}  \mathfrak{M}_t 
\leq \epsilon^2$, Let B = T, and we choose T $\geq \epsilon^{-2}$, the total samples evaluated across the network system is on the order of $O(\epsilon^{-4})$.
\end{remark}

\begin{figure*}[ht]  
\centering
\subfigure[MNIST]{
\hspace{0pt}
\includegraphics[width=.31\textwidth]{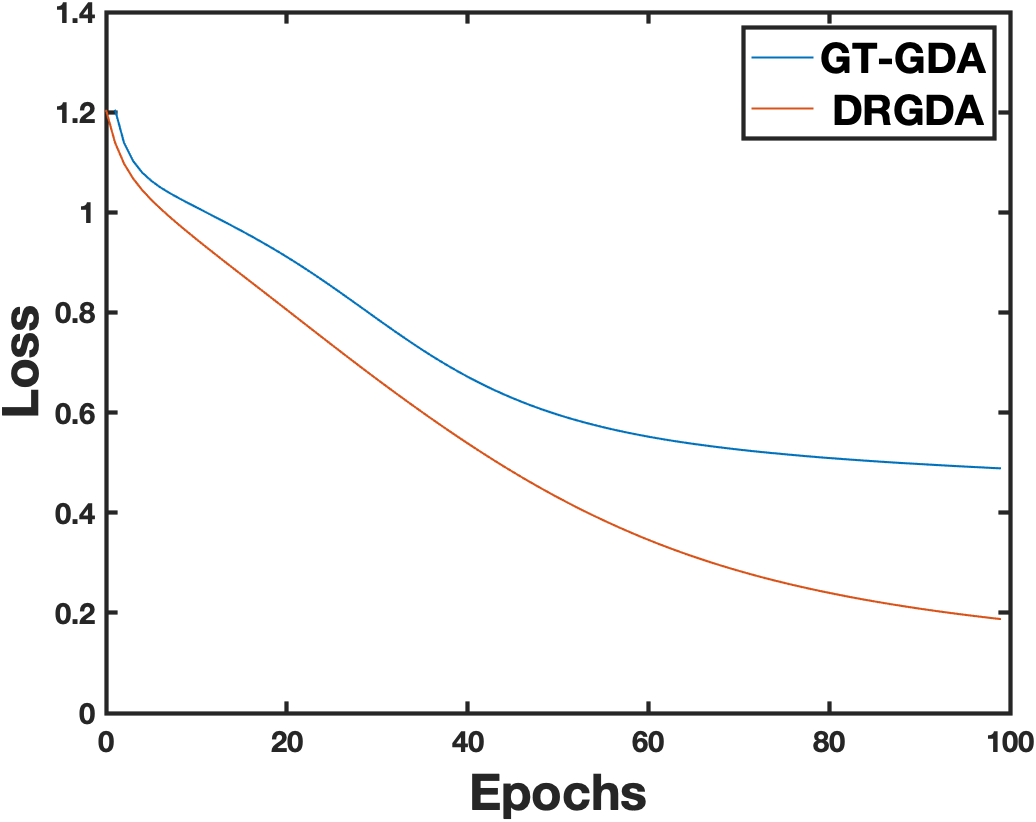}
}
\subfigure[Fashion-MNIST]{
\hspace{0pt}
\includegraphics[width=.31\textwidth]{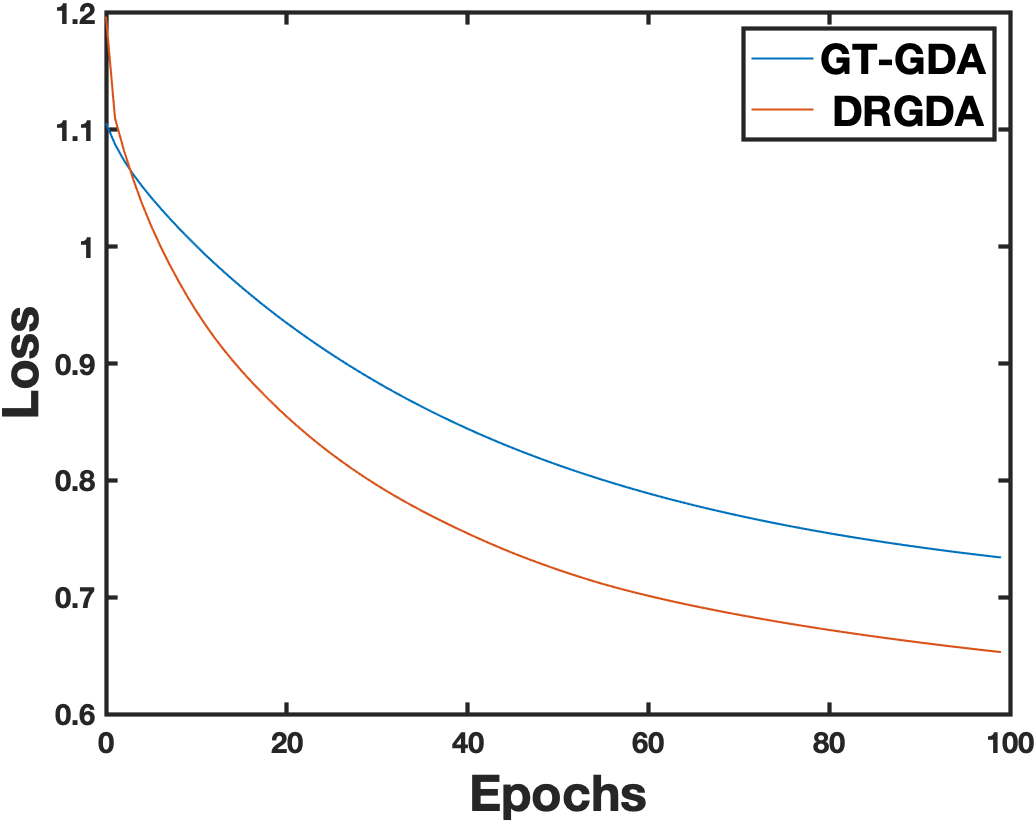}
}
\subfigure[CIFAR-10]{
\hspace{0pt}
\includegraphics[width=.31\textwidth]{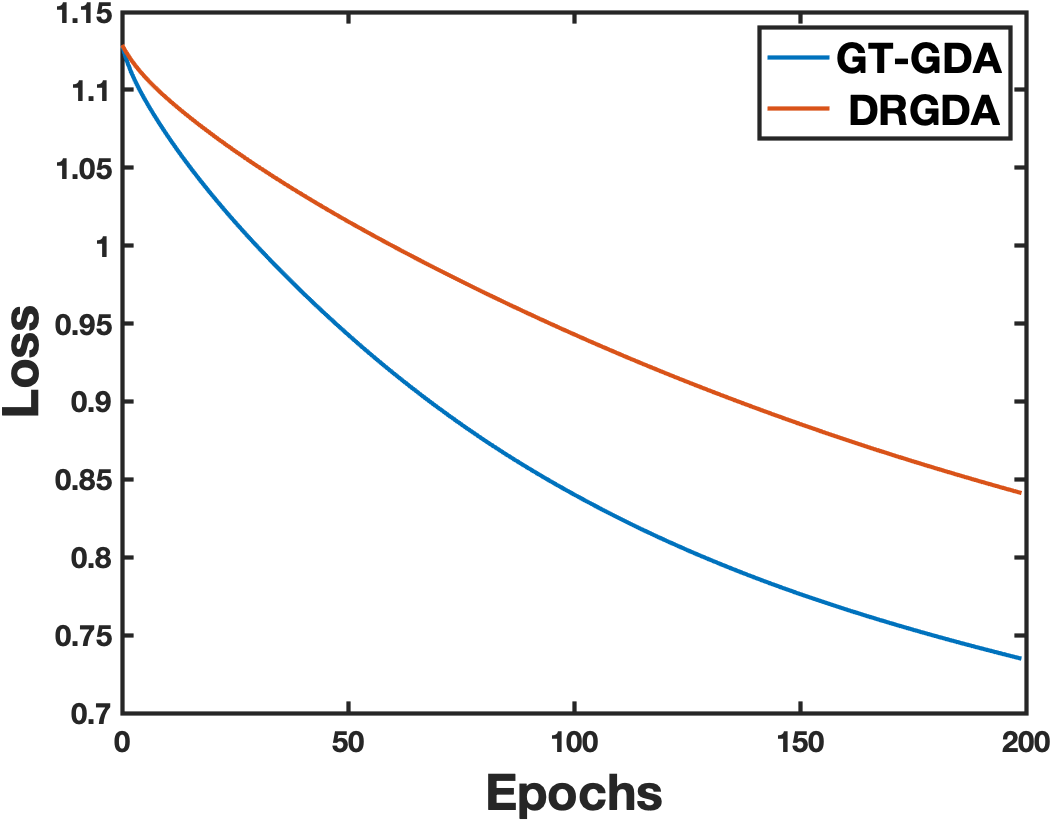}
}
\caption{Results of different deterministic methods on the orthonormal fair classification networks task.}
\label{img:1}
\end{figure*}

\begin{figure*}[ht]
\centering
\subfigure[MNIST]{
\hspace{0pt}
\includegraphics[width=.31\textwidth]{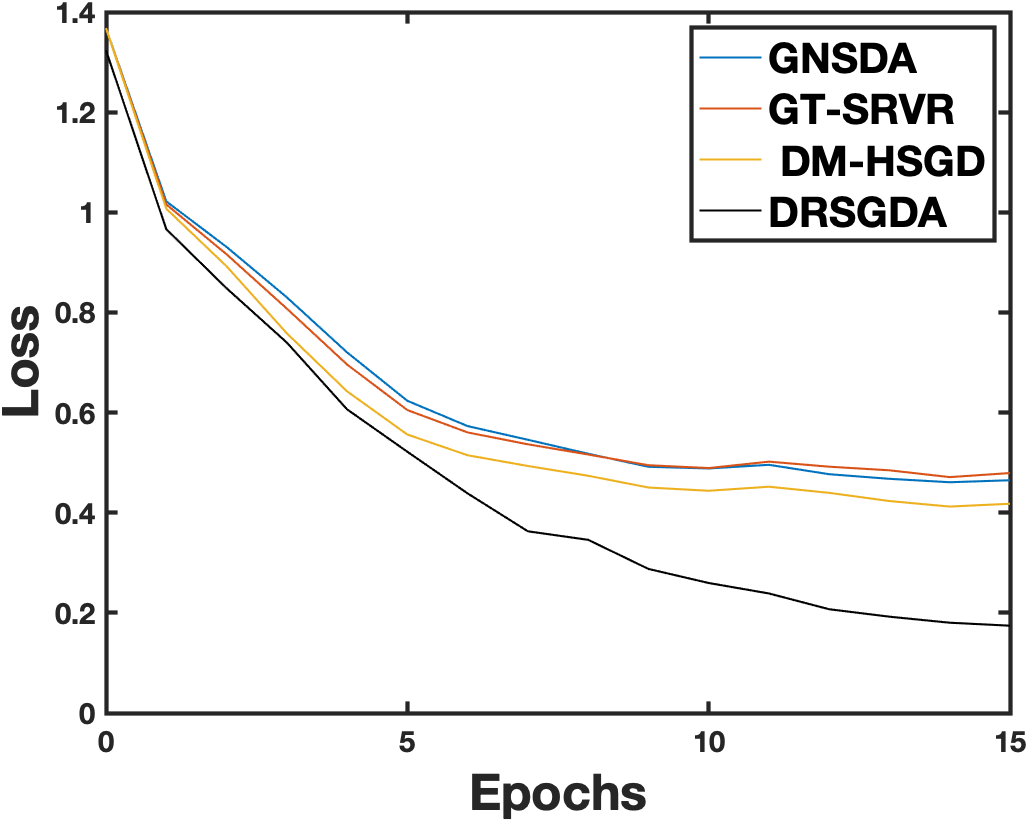}
}
\subfigure[Fashion-MNIST]{
\hspace{0pt}
\includegraphics[width=.31\textwidth]{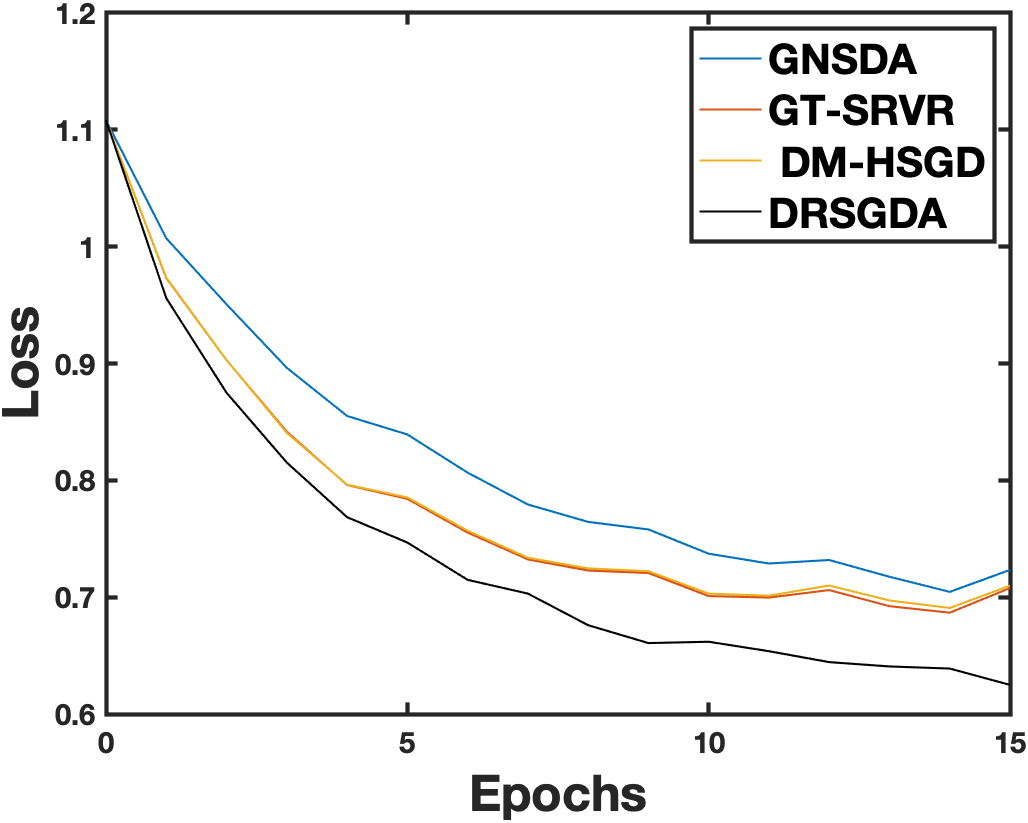}
}
\subfigure[CIFAR-10]{
\hspace{0pt}
\includegraphics[width=.31\textwidth]{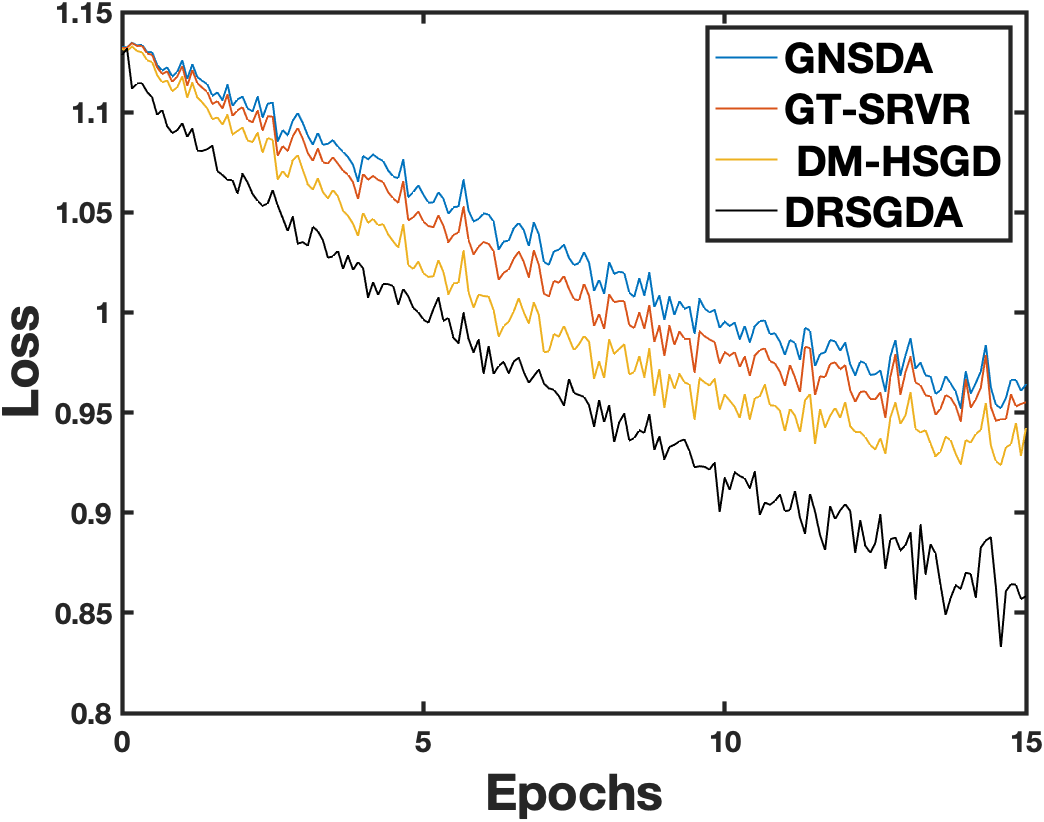}
}
\caption{Results of different stochastic methods on the orthonormal fair classification networks task.}
\label{img:2}
\end{figure*}

\section{Numerical Experiments}
We conducted numerical experiments to validate the efficiency of our algorithms on two tasks: 1) Orthonormal fair classification networks and 2) distributionally robust optimization with orthonormal weights.  
In the deterministic setting, we use the GT-GDA \citep{zhang2021taming} as the baseline to compare with DRDGA. In the stochastic setting, we use GNSDA which is motivated by the GNSD algorithm \citep{lu2019gnsd}, DM-HSGD \citep{xian2021faster},  GT-SRVR \citep{zhang2021taming} as the comparison baselines for solving nonconvex-strongly-concave minimax problems. 
Since these methods were not designed for optimization on the Stiefel manifold, we add the retraction operation (projection-like) when we do the test experiments. The experiments are conducted using computers with 2.3 GHz Intel Core i9 CPUs and NVIDIA Tesla P40 GPUs. We set the number of worker nodes as $n = 20$ and use the ring-based topology as the communication network \citep{xian2021faster}. 

\subsection{Orthonormal Fair Classification Networks}
In the first task, we train orthonormal fair classification networks by minimizing the maximum loss over different categories. Orthonormality on parameters corresponds to optimization over the Stiefel manifold. We use Convolutional Neural Networks (CNNs) as classifiers, whose architectures are given in supplementary materials. In the experiment, we use the MNIST, Fashion-MNIST, and CIFAR-10 datasets as in \citep{huang2021efficient}. Following \citep{huang2021efficient}, we mainly focus on three categories in each dataset: digital numbers $\{0, 2, 3\}$ in the MNIST dataset, and T-shirt/top, Coat, and Shirt categories in the Fashion-MNIST dataset, and airplane, automobile, and bird in the CIFAR10 dataset. The datasets are evenly divided into disjoint sets across all worker nodes. Then we train this fair classifier by solving the following minimax problem:
\begin{align} \label{eq:28}
\min_{w \in \mathrm{St}(d,r)} \max \{\mathcal{L}_1(w), \mathcal{L}_2(w), \mathcal{L}_3(w) \} \,,
\end{align}
where $\mathcal{L}_{1}$, $\mathcal{L}_{2}$, and $\mathcal{L}_{3}$ are the cross-entropy loss functions corresponding to the samples in three different categories and $w$ denotes the CNN model parameters. Problem \eqref{eq:28} can be re-written as the following:
\begin{align}
  & \min_{w \in \mathrm{St}(d,r)} \max_{ u\in \mathcal{U} } \ \big\{\sum_{i=1}^{3} u_i \mathcal{L}_i(w)- \rho\|u\|^2  \big\}  \\   \label{eq:23}
  &  \text{s.t.}  \quad \mathcal{U}= \big\{ u \ | \  u_i \geq 0, \ \sum_{i=1}^{3} u_{i}=1 \big\}\,, \nonumber 
\end{align}
where $\rho>0$ is the tuning parameter, and $u$ is a weight vector for different loss functions. 

In the experiment, we use Xavier normal initialization to CNN  layer. The grid search is used to tune parameters for all methods. For all datasets, we choose the $\{\alpha, \beta, \eta \}$ from the set $\{0.0001, 0.001, 0.005, 0.01\}$ for DRGDA and DRSGDA. For other methods, we tune the learning rates from the set $\{0.0001, 0.001, 0.005, 0.01\}$. For DM-HSGD, we also set $\{\beta_x, \beta_y\}$ from the set $\{0.1, 0.9 \}$. The batch sizes for MNIST and Fashion-MNIST are 100 while that for CIFAT-10 is 50. The initial batch sizes for GT-SRVR and DM-HSGD are set as 300. 

The training progress for orthonormal fair classification networks is reported in Figures \ref{img:1} and \ref{img:2}. We can see that our methods (DRGDA and DRSGDA) converge faster than the other comparison baselines on all datasets.

\subsection{Distributionally Robust Optimization}
In the second task, we focus on distributionally robust optimization with orthonormal weights which is defined as the following:
\begin{align}
\min _{w \in \mathrm{St}(d,r)} \max_{\mathbf{p} \in \Delta_n}\left\{\sum_{i=1}^{n} p_{i} \ell_i \left(w ; \xi_{i}\right)-\left\|\mathbf{p}-\frac{1}{n}\right\|^{2}\right\}    
\end{align}
where $\mathbf{p} = (p_1, \cdots, p_n)$ is the i-th component of variable y. $\Delta_n$ represents the simplex in $\mathbb{R}^n$.  $\ell_i(w;\xi)$ denotes the loss function over the Stiefel manifold.

In this task, we also use the datasets MNIST, Fashion-MNIST, and CIFAR-10 datasets with the same DNN architecture provided in the supplementary materials. Similarly, the datasets are evenly divided by all worker nodes. We also use the grid search to tune parameters. For all methods, we tune the step size from the set $\{0.001, 0.01, 0.1, 0.5, 1\}$. For DM-HSGD, we set $\{\beta_x, \beta_y\}$ from the set $\{0.1, 0.9 \}$. The batch size for all methods is set as 100. The initial batch size for GT-SRVR and DM-HSGD is set as 300. 

The experimental results are reported in supplementary materials. From the results, we can see that our method DRSGDA achieves the best test accuracy and converges fastest compared with other baseline methods.

\section{Conclusion}
In the paper, we proposed decentralized Riemannian gradient descent ascent methods for solving the decentralized Riemannian minimax problems \eqref{eq:1} and \eqref{eq:2}, respectively. We focus on non-convex strong-concave and $x \in \mathrm{St}(d,r)$. In particular, our methods are the first algorithms for distributed minimax optimization with nonconvex constraints with exact convergence. For the deterministic setting, we introduced the decentralized gradient descent ascent algorithm (DRGDA) over the Stiefel manifold. We proved that the DRGDA has a sample complexity of $O(\epsilon^{-2})$ for finding an $\epsilon$-stationary point, which matches the best complexity result achieved by the centralized minimax method on the Euclidean space. We also propose a decentralized stochastic first-order algorithm (DRSGDA) over the Stiefel manifold for the stochastic setting, which has a sample complexity of $O(\epsilon^{-4})$. This is the first theoretical result of the stochastic method in the decentralized Riemannian minimax problems.

\section*{Acknowledgements}
This work was partially supported by NSF IIS 1838627, 1837956, 1956002, 2211492, CNS 2213701, CCF 2217003, DBI 2225775.

\bibliography{aaai23}


\end{document}